%% file: human.tex
\documentclass[10pt,twocolumn,letterpaper]{article}
\usepackage{cvpr}
\usepackage{times}
\usepackage{epsfig}
\usepackage{graphicx}
\usepackage{amsmath}
\usepackage{amssymb}
\usepackage{bbm}
\usepackage{multirow}
\usepackage{makecell}
\usepackage{color}
\usepackage{caption}
\usepackage{amsmath}
\newcommand{\bm}[1]{\mathbf{#1}}
\newcommand{\myb}{\mbox{\boldmath $\beta$}}
\newcommand{\myt}{\mbox{\boldmath $\theta$}}
\newcommand{\T}{\Theta}
\newcommand{\delT}{\Delta\Theta}
\newcommand{\blue}[1]{\textcolor{black}{#1}}
\usepackage[pagebackref=true,breaklinks=true,letterpaper=true,colorlinks,bookmarks=false]{hyperref}

\cvprfinalcopy %*** Uncomment this line for the final submission

\ifcvprfinal\fi
\begin{document}

%%%%%%%%% TITLE
\title{End-to-end Recovery of Human Shape and Pose}
% The \author macro works with any number of authors. There are two
% commands used to separate the names and addresses of multiple
% authors: \And and \AND.
%
% Using \And between authors leaves it to LaTeX to determine where to
% break the lines. Using \AND forces a line break at that point. So,
% if LaTeX puts 3 of 4 authors names on the first line, and the last
% on the second line, try using \AND instead of \And before the third
% author name.

\author{Angjoo Kanazawa$^{1}$, Michael J. Black$^2$, David W. Jacobs$^3$, Jitendra
  Malik$^1$\\
  $^1$University of California, Berkeley\\
$^2$MPI for Intelligent Systems, T\"{u}bingen, Germany, $^3$University of Maryland, College Park\\
{\tt\small \{kanazawa,malik\}@eecs.berkeley.edu},
{\tt\small black@tuebingen.mpg.de},
{\tt\small djacobs@umiacs.umd.edu}}

\twocolumn[{%
\renewcommand\twocolumn[1][]{#1}%
\maketitle
\begin{center}
    \newcommand{\teaserwidth}{\textwidth}
% \vspace{-0.35in}
    \centerline{
   \includegraphics[width=\teaserwidth,clip]{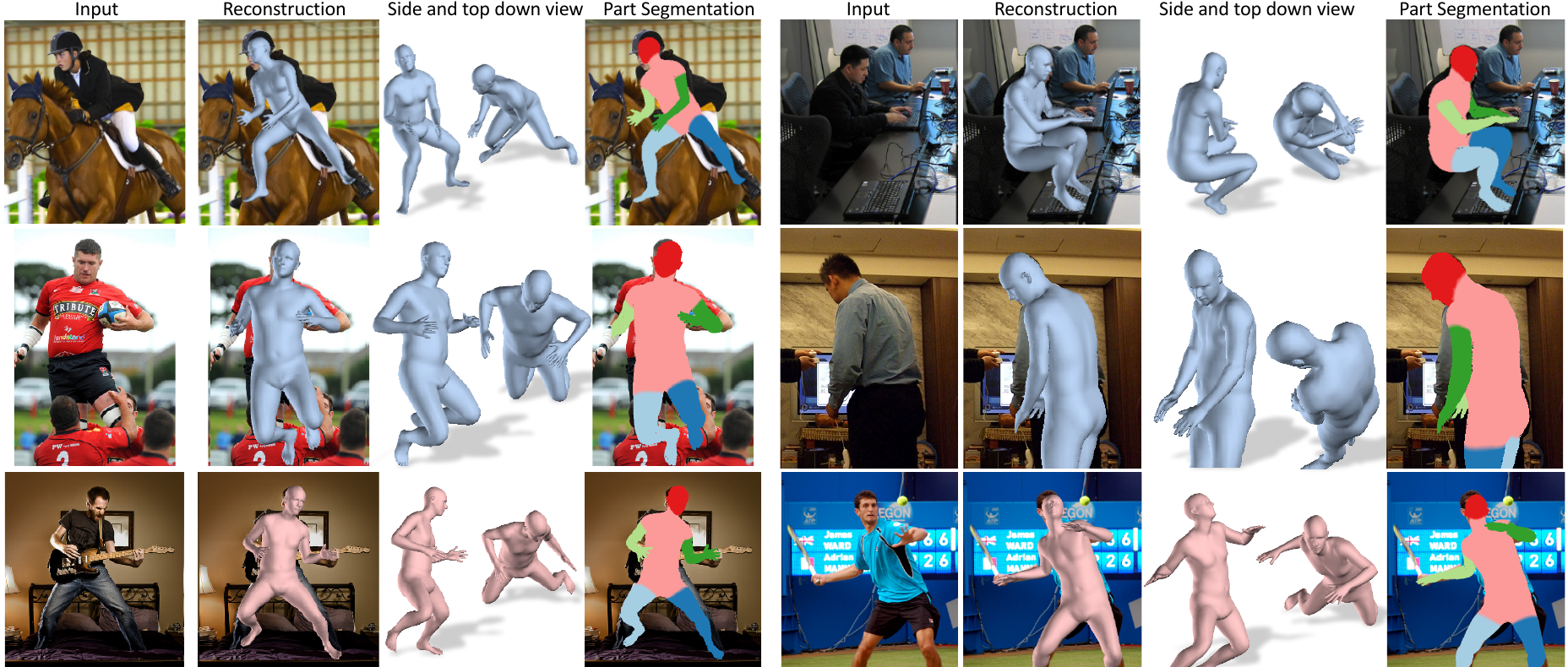}
     }
   \captionof{figure}{{\bf \blue{Human Mesh Recovery (HMR):} End-to-end adversarial
       learning of human pose and shape}. We describe a real time framework for recovering
     the 3D joint angles and shape of the body from a single RGB image. The
     first two rows show results from our model trained with some 2D-to-3D
     supervision, the bottom row shows results from a model that is trained in a fully weakly-supervised
     manner without using any paired 2D-to-3D
     supervision. We infer the full 3D body even in case of occlusions and
     truncations. Note that we capture head and limb orientations.}
   \vspace{-0.05in}
\label{fig:teaser}
\end{center}%
}]
% \maketitle

\input{src/abstract}
\input{src/intro}
\input{src/related}

\input{src/model}
\input{src/detail}

\input{src/experiments}
\input{src/discussion}

\small
\bibliographystyle{ieee}
\bibliography{human}

\end{document}

%% file: src/abstract.tex
\begin{abstract}
\vspace{-0.1in}  
We describe Human Mesh Recovery (HMR), an end-to-end framework for reconstructing a full
3D mesh of a human body from a single RGB image. 
In contrast to most current methods that compute 2D or 3D joint
locations, we produce a richer and more useful mesh representation that is
parameterized by shape and 3D joint angles. The main objective is to minimize
the reprojection loss of keypoints, which allows our model to be trained using \emph{in-the-wild} images that only have
ground truth 2D annotations.
However, the reprojection loss alone is highly underconstrained.
In this work we address this problem by introducing an adversary trained to
tell whether human body shape and pose parameters are real or not using a large database of
3D human meshes. We show that HMR can be trained with and \emph{without} using any paired 2D-to-3D supervision. 
We do not rely on intermediate 2D keypoint detections and infer 3D pose and shape
parameters directly from image pixels. Our model runs in real-time given a bounding box containing the person.  We demonstrate
our approach on various images \emph{in-the-wild} and out-perform previous
optimization-based methods that output 3D meshes and show competitive results on tasks such as 3D joint location estimation and part segmentation.
\end{abstract}

%% file: src/intro.tex
\vspace{-0.1in}
\begin{figure*}[h]
  \centering
  \includegraphics[width=\textwidth]{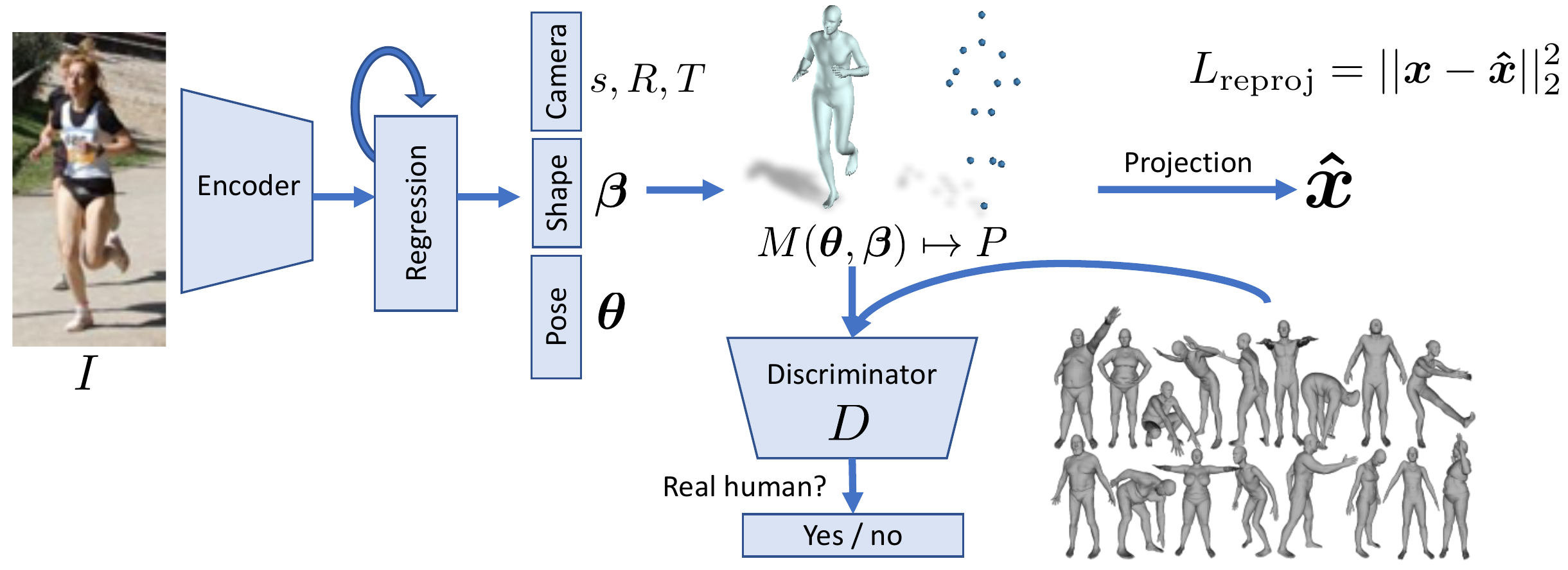}
  \caption{{\bf Overview of the proposed framework.} 
    An image $I$ is passed through a convolutional encoder. This is sent to an iterative 3D regression
    module that infers the latent 3D representation of the human that minimizes
    the joint reprojection error. The 3D parameters are also sent to the
    discriminator $D$, whose goal is to tell if these parameters come from a real human shape and pose.
  }
  % \vspace{-.5em}
  \label{fig:overview}
\end{figure*}
\section{Introduction}
\label{sec:intro}
We present an end-to-end framework for recovering a full 3D mesh of a human body
from a single RGB image. We use the generative human
body model, SMPL \cite{SMPL}, which parameterizes the mesh by 3D joint
angles and a low-dimensional linear shape space.
As illustrated in Figure \ref{fig:teaser}, estimating a 3D mesh opens the door
to a wide range of applications such as foreground and part segmentation, which
is beyond what is practical with a simple skeleton.
The output mesh can be immediately used by animators, modified, measured, manipulated and retargeted.
Our output is also holistic -- we always infer the full 3D body even in
cases of occlusion and truncation.

Note that there is a great deal of work on the 3D analysis of humans from a single
image. 
Most approaches, however, focus on recovering 3D joint
locations. We argue that these joints alone are not the full story. 
Joints are sparse, whereas the human body is defined by a surface in
3D space.

Additionally, joint locations alone do not constrain the full DoF at each joint.
This means that it is non-trivial to estimate the full pose of the
body from only the 3D joint locations.
In contrast, we output the relative 3D rotation matrices for each
joint in the kinematic tree, capturing information about
3D head and limb orientation. Predicting rotations also ensures that limbs are symmetric and of valid length. Our model implicitly learns the
joint angle limits from datasets of 3D body models. 

% Goal:
Existing methods for recovering 3D human mesh today focus on a multi-stage
approach \cite{SMPLify,UP}.
First they estimate 2D joint locations and, from these, estimate the 3D model parameters.
Such a stepwise approach is typically not optimal and here we propose
an end-to-end solution to learn a mapping from image pixels directly to model parameters.

%Problems:
There are several challenges, however, in training such a model in an end-to-end
manner. First is the lack of large-scale ground truth 3D annotation for \emph{in-the-wild}
images. Existing datasets with accurate 3D annotations are captured in
constrained environments. Models trained on these datasets do not generalize well
to the richness of images in the real world. 
Another challenge is in the inherent ambiguities in single-view 2D-to-3D
mapping. Most well known is the problem of depth ambiguity where multiple 3D
body configurations explain the same 2D projections \cite{Taylor:2000}. Many of these configurations may not be
anthropometrically reasonable, such as impossible joint angles or extremely
skinny bodies. In addition, estimating the camera explicitly introduces an
additional scale ambiguity between the size of the person and the camera distance.

%Idea/solution
In this paper we propose a novel approach to mesh reconstruction that addresses
both of these challenges. 
A key insight is that there are large-scale 2D
keypoint annotations of \emph{in-the-wild} images and a separate
large-scale dataset of 3D meshes of people with various poses and shapes. 
Our key contribution is to take advantage of these
\emph{unpaired} 2D keypoint annotations and 3D scans in a conditional
{\em generative adversarial manner.}
The idea is that, given an image, the network has to infer the 3D mesh parameters
and the camera such that % , when projected into the image,
the 3D keypoints match the annotated 2D keypoints after projection. To deal with ambiguities, these parameters are sent to a
discriminator network, whose task is to determine if the 3D parameters correspond to bodies of real humans or not. 
Hence the network is encouraged to output parameters on the human
manifold and the discriminator acts as weak supervision. % data-driven prior.
The network implicitly learns the angle limits for each joint and is
discouraged from making people with unusual body shapes. 

An additional challenge in predicting body model parameters is that
regressing to rotation matrices is challenging. 
Most approaches formulate rotation estimation as a classification problem
by dividing the angles into bins \cite{VPKP}. However differentiating angle
probabilities with respect to the reprojection loss is non-trivial and
discretization sacrifices precision. Instead we propose to directly regress these values in an iterative manner with feedback. Our framework is illustrated in Figure \ref{fig:overview}.

Our approach is similar to 3D interpreter networks \cite{Pavlakos,3dinterpreter} in the use of reprojection
loss and the more recent adversarial inverse graphics networks \cite{Fish2017}
for the use of the adversarial prior.
We go beyond the existing techniques in multiple ways:
\begin{enumerate}
\item We infer 3D mesh parameters directly from
image features, while previous approaches infer them from 2D keypoints. This
avoids the need for two stage training and also avoids throwing away valuable
information in the image such as context.

\item Going beyond skeletons, we output meshes, which are more complex 
  and more appropriate for many applications. Again, no additional
  inference step is needed.
\item Our framework is trained in an end-to-end manner. We out-perform
  previous approaches that output 3D meshes \cite{SMPLify,UP} in terms of 3D
  joint error and run time.
\item We show results with and \emph{without} paired 2D-to-3D data. Even without
  using any paired 2D-to-3D supervision, our approach produces reasonable 3D
  reconstructions. This is most exciting because it opens up possibilities for
  learning 3D from large amounts of 2D data.
\end{enumerate}

Since there are no datasets for evaluating 3D mesh reconstructions
of humans from in-the-wild images, we are bound to evaluate our
approach on the standard 3D joint location estimation task. Our
approach out performs previous methods that estimate SMPL parameters from 2D
joints and is competitive with approaches that only output 3D skeletons. We
also evaluate our approach on an auxiliary task of human part segmentation. We
qualitatively evaluate our approach on challenging images in-the-wild and show
results sampled at different error percentiles. Our model and code
is available for research purposes at \url{https://akanazawa.github.io/hmr/}.

%% file: src/related.tex
\section{Related Work}
\label{sec:related}
\paragraph{3D Pose Estimation:} 
Many papers formulate human pose estimation as the problem of locating
the major 3D joints of the body from an image, a video sequence, either single-view or
multi-view. 
We argue that this notion of ``pose'' is overly simplistic but it is
the major paradigm in the field. 
The approaches are split into two categories: two-stage and direct estimation.

Two stage methods first predict 2D joint locations using 2D pose detectors 
\cite{hourglass,Wei:CVPR:2016,Zhou:2016a} or ground truth 2D pose and
then predict 3D joint locations from
the 2D joints either by regression \cite{Martinez2017,moreno20163d} or model fitting,
where a common approach exploits a learned dictionary of 3D skeletons \cite{Akhter, ramakrishna2012reconstructing,Fish2017,Sanzari2016,Zhou:2015b,Zhou:2016a}.
In order to constrain the inherent ambiguity in 2D-to-3D estimation, these
methods use various priors \cite{Taylor:2000}. Most
methods make some assumption about the limb-length or proportions
\cite{BarronKakadiaris2001,Lee1985,Parameswaran:CVPR:2004,ramakrishna2012reconstructing}. Akhter
and Black \cite{Akhter} learn a novel pose prior that captures
pose-dependent joint angle limits. Two stage-methods have the benefit of being
more robust to domain shift, but rely too much on 2D joint detections
and may throw away image information in estimating 3D pose. 

Video datasets with ground truth motion capture like HumanEva
\cite{HumanEva:2010} and Human3.6M \cite{Human36m:2014} define the
problem in terms of 3D joint locations. 
They provide training data that lets the 3D joint estimation
problem be formulated as a standard supervised learning problem. 
Thus, many recent methods estimate 3D joints directly from images in a deep
learning framework \cite{Pavlakos,Tekin2017,Tome_2017_CVPR,Xingyi2017,Xingyi2016}. Dominant approaches are
fully-convolutional, except for the very recent method of Xiao \etal
\cite{Xiao2017} that regresses bones and obtains excellent results on the 3D
pose benchmarks. Many methods do not solve for the camera, but estimate the depth relative
to root and use a predefined global scale based the average length of bones
\cite{Pavlakos,Xingyi2017,Xingyi2016}. Recently Rogez \etal \cite{Rogez} combine human detection
with 3D pose prediction. The main issue with these direct estimation methods is
that images with accurate ground truth 3D annotations are captured in
controlled MoCap environments. Models trained only on these images do
not generalize well to
the real world.

\noindent {\bf Weakly-supervised 3D:}
Recent work tackles this problem of the domain gap between
MoCap and \emph{in-the-wild} images in an end-to-end framework. Rogez and Schmid
\cite{RogezMocap} artificially endow 3D annotations to images with 2D pose annotation using
MoCap data. 
Several methods \cite{Mehta17,VNect,Xingyi2017} train on both in-the-wild and MoCap
datasets jointly. 
Still others \cite{Mehta17,VNect} use pre-trained 2D pose networks and also use 
2D pose prediction as an auxiliary task.
When 3D annotation is not available, Zhou \etal \cite{Xingyi2017} gain weak
supervision from a geometric constraint that encourages relative bone lengths to
stay constant. In this work, we output 3D joint angles and 3D shape, which subsumes
these constraints that the limbs should be symmetric. We employ a much stronger
form of weak supervision by training an adversarial prior.

\noindent{\bf Methods that output more than 3D joints:}
There are multiple methods that fit a parametric body model to manually extracted
silhouettes \cite{Chen:ECCV:2010} and a few manually provided correspondences
\cite{ Guan:2009,Hasler:2010}. More recent works attempt to automate this
effort. Bogo \etal \cite{SMPLify} propose SMPLify, an optimization-based method to
recover SMPL parameters from 14 detected 2D joints that leverages multiple
priors. 
However, due to the optimization steps the approach is not real-time,
requiring 20-60 seconds per image. They also make \emph{a priori} assumptions about
the joint angle limits. 
Lassner \etal \cite{UP} take curated results from SMPLify to train 91
keypoint detectors corresponding to traditional body joints and points on the surface.
They then optimize the SMPL model parameters to fit the keypoints
similarly to \cite{SMPLify}.
They also propose a random forest regression approach to directly regress SMPL parameters, which reduces run-time at the cost
of accuracy. 
Our approach out-performs both methods, directly infers SMPL parameters
from images instead of detected 2D keypoints, and runs in real time.

VNect \cite{VNect} fits a rigged skeleton model over time to estimated
2D and 3D joint locations.
While they can recover 3D rotations of each joint after optimization, we directly output rotations from
images as well as the surface vertices. Similarly Zhou \etal \cite{Xingyi2016} directly
regress joint rotations of a fixed kinematic tree. We output shape as well as
the camera scale and out-perform their approach in 3D pose estimation.

There are other related methods that predict SMPL-related outputs:
Varol \etal \cite{Gul} use a synthetic
dataset of rendered SMPL bodies to learn a fully convolutional model
for depth and body part segmentation. 
DenseReg \cite{DenseReg} similarly outputs a dense
correspondence map for human bodies. Both are 2.5D projections of the
underlying 3D body. In this work, we recover all SMPL parameters
and the camera, from which all of these outputs can be obtained.

Kulkarni \etal~\cite{Kulkarni_2015_CVPR} use a generative model of body shape
and pose with a probabilistic programming framework to
estimate body pose from single image. They deal with visually simple
images and do not evaluate 3D pose accuracy. More recently Tan \etal \cite{Tan} infer SMPL 
parameters by first learning a silhouette decoder of SMPL
parameters using synthetic data, and then learning an image encoder with the
 decoder fixed to minimize the silhouette reprojection loss. However, the reliance on
 silhouettes limits their approach to frontal images and images of humans without any occlusion. 
Concurrently Tung \etal \cite{tung2017self} predict SMPL
parameters from an image and a set of 2D joint heatmaps. The model is pretrained on a synthetic dataset and fine-tuned at test
time over two consecutive video frames to minimize the reprojection loss of keypoints, silhouettes and optical
flow. Our approach can be trained without any paired
supervision, does not require 2D joint heatmaps as an input and we test on
images without fine-tuning. Additionally, we also demonstrate our approach
on images of humans in-the-wild
\cite{coco} with clutter and occlusion.

%% file: src/model.tex
\section{Model}
We propose to reconstruct a full 3D mesh of a human body directly from a single RGB
image $I$ centered on a human in a feedforward manner. During training we assume that all
images are annotated with ground truth 2D joints.
We also consider the case in which some have 3D annotations
as well. 
Additionally we assume that there is a pool of 3D meshes of
human bodies of varying shape and pose. 
Since these meshes do not necessarily have a corresponding image,
we refer to this data as \emph{unpaired} \cite{CycleGAN2017}.

Figure \ref{fig:overview} shows the overview of the proposed network
architecture, which can be trained end-to-end. Convolutional
features of the image are sent to the iterative 3D regression
module whose objective is to infer the 3D human body and the camera such that
its 3D joints \emph{project} onto the annotated 2D joints. 
The inferred parameters are also sent to an adversarial discriminator network
whose task is to determine if the 3D parameters are real meshes from the
\emph{unpaired} data. This encourages the network to output 3D human
bodies that lie on the manifold of human bodies and acts as a weak-supervision for
\emph{in-the-wild} images without ground truth 3D annotations. Due to the rich
representation of the 3D mesh model, this data-driven prior can capture joint angle
limits, anthropometric constraints (\eg height, weight, bone ratios), and subsumes the geometric
priors used by models that only predict 3D joint locations \cite{ramakrishna2012reconstructing,Xiao2017,Xingyi2017}. 
When ground truth 3D information is available, we may use it as an intermediate
loss. In all, our overall objective is 
\begin{equation}
  \label{eq:overall_loss}
  L = \lambda(L_{\text{reproj}} + \mathbbm{1}L_{\text{3D}}) + L_{\text{adv}}
\end{equation}
where $\lambda$ controls the relative importance of each objective, $\mathbbm{1}$ is an indicator function that is 1 if ground truth 3D is available
for an image and 0 otherwise. We show results with and \emph{without} the 3D
loss. We discuss each component in the following.

\subsection{3D Body Representation}
We encode the 3D mesh of a human body using the  Skinned Multi-Person
Linear (SMPL) model \cite{SMPL}. SMPL is a
generative model that factors human bodies into \emph{shape} -- how individuals vary in height, weight, body proportions -- and \emph{pose} -- how the 3D
surface deforms with articulation. The shape $\myb \in \mathbb{R}^{10}$ is
parameterized by the first 10 coefficients of a PCA shape space. The pose $\myt \in \mathbb{R}^{3K}$ is modeled by relative
3D rotation of $K=23$ joints in axis-angle representation. SMPL is a differentiable function that outputs a triangulated mesh with $N=6980$
vertices, $M(\myt, \myb) \in \mathbb{R}^{3\times N}$, which is obtained by
shaping the template body vertices conditioned on $\myb$ and $\myt$, then
articulating the bones according to the joint rotations $\myt$ via forward kinematics, and finally
deforming the surface with linear blend skinning. 
The 3D keypoints used for reprojection error, $X(\myt, \myb) \in \mathbb{R}^{3
  \times P}$, are obtained by linear regression from the final mesh vertices.

We employ the weak-perspective camera model and solve for the global
rotation $R\in \mathbb{R}^{3\times 3}$ in axis-angle representation, translation $t
\in \mathbb{R}^2$ and scale $s \in \mathbb{R}$. Thus the set of parameters that represent the 3D reconstruction of a human
body is expressed as a 85 dimensional vector $\Theta = \{\myt, \myb, R, t, s\}$. Given
$\T$, the projection of $X(\myt, \myb)$ is
\begin{equation}
  \label{eq:proj}
\hat{\bm{x}} = s\Pi (RX({\myt, \myb})) + t,
\end{equation}
where $\Pi$ is an orthographic projection.

\subsection{Iterative 3D Regression with Feedback}
The goal of the 3D regression module is to output $\T$ given an image encoding $\phi$ such that the joint reprojection error
\begin{equation}
  \label{eq:reproj}
  L_{\text{reproj}} = \Sigma_i ||v_i(\bm{x}_i - \hat{\bm{x}}_i) ||_1,
\end{equation}
 is minimized. Here $\bm{x}_i \in \mathbb{R}^{2 \times K}$ is the $i$th ground truth 2D joints and
 $v_i \in \{0,1\}^{K}$ is the visibility (1 if visible, 0 otherwise) for each
of the $K$ joints.

However, directly regressing $\T$ in one go is a challenging task, particularly
because $\T$ includes rotation parameters.
In this work, we take inspiration from previous works
\cite{IEF,dollar2010cascaded,oberweger2015training} and regress $\T$ in an iterative error feedback
(IEF) loop, where progressive changes are made recurrently to the current estimate. Specifically,
the 3D regression module takes the image features $\phi$ and the current
parameters $\T_t$ as an input and outputs the residual $\delT_t$. The parameter
is updated by adding this residual to the current estimate $\T_{t+1} = \T_t +
\delT_t$. The initial estimate $\T_{0}$ is set as the mean $\bar \T$. 
In \cite{IEF,oberweger2015training} the estimates are rendered to an image space to
concatenate with the image input. In this work, we keep everything in the latent
space and simply concatenate the features $[\phi, \T]$ as the input to the
regressor. We find that this works well and is suitable when differentiable
rendering of the parameters is non-trivial.

Additional direct 3D supervision may be employed when paired ground truth 3D
data is available. The most common form of 3D annotation is the 3D
joints. Supervision in terms of SMPL parameters $[\myb, \myt]$ may be obtained
through MoSh \cite{Mosh,Gul} when raw 3D MoCap marker data is available. Below
are the definitions of the 3D losses. We show results with and without using any direct supervision $L_{\text{3D}}$. 
\begin{align}
  \label{eq:3dloss}
  L_{\text{3D}} &= L_{\text{3D joints}} + L_{\text{3D smpl}}\\
  L_{\text{joints}} &= ||(\bm{X_i} - \hat{\bm{X_i}}) ||_2^2 \\
  L_{\text{smpl}} &= ||[\myb_i, \myt_i] - [\hat{\myb_i},\hat{\myt_i}]||_2^2.
\end{align}

Both \cite{IEF,oberweger2015training} use a ``bounded'' correction target to
supervise the regression output at each iteration. However this assumes that the ground truth estimate is always
known, which is not the case in our setup where many images do not have ground truth 3D annotations.
As noted by these approaches, supervising each iteration with the final objective forces
the regressor to overshoot and get stuck in local minima. Thus we only apply
$L_{\text{reproj}}$ and $L_{\text{3D}}$ on the final estimate $\Theta_T$, but apply the adversarial loss on the
estimate at every iteration $\Theta_t$ forcing the network to take corrective steps that
are on the manifold of 3D human bodies. 

\subsection{Factorized Adversarial Prior}
The reprojection loss encourages the network to produce a 3D body that explains the
2D joint locations, however anthropometrically implausible 3D bodies or bodies with gross self-intersections may still minimize the reprojection loss.
To regularize this, we use a discriminator network $D$ that is trained to tell
whether SMPL parameters correspond to a real body or not. We refer to this as an
adversarial prior as in \cite{Fish2017} since the discriminator acts as a
data-driven prior that guides the 3D inference. 

A further benefit of employing a rich, explicit 3D representation like SMPL is that we precisely know
the meaning of the latent space. In particular SMPL has a factorized form that
we can take advantage of to make the adversary \blue{more data efficient and stable} to train. More concretely, we mirror the shape and pose decomposition of SMPL
and train a discriminator for shape and pose independently. 
The pose is based on a kinematic tree, so we further decompose the pose
discriminators and train one for each joint rotation. This amounts to learning the angle limits for each
joint. In order to capture the joint distribution of the entire kinematic tree,
we also learn a discriminator that takes in all the rotations.
Since the input to each discriminator is very low dimensional (10-D for
$\myb$, 9-D for each joint and $9K$-D for all joints), they can each be small
networks, making them rather stable to train. All pose discriminators share a
common feature space of rotation matrices and only the final classifiers are learned
separately. 

Unlike previous approaches that make \emph{a priori} assumptions about the joint
limits \cite{SMPLify,Xingyi2016}, we do not predefine the degrees of freedom of
the kinematic skeleton model. Instead this is learned in a data-driven
manner through this factorized adversarial prior. Without the factorization, the
network does not learn to properly regularize the pose and shape, producing
visually displeasing results. 
The importance of the adversarial prior is paramount when no paired 3D supervision is available. Without the adversarial
prior the network produces totally unconstrained human bodies as we show in
section \ref{sec:no3d}. 

While mode collapse is a common issue in GANs
\cite{goodfellow2014generative} we do not really suffer from this because the
network not only has to fool the discriminator but also has to minimize the
reprojection error. The images contain all the modes and the network is forced
to match them all. The factorization may further help to avoid mode collapse
since it allows generalization to unseen body shape and poses combinations. 

In all we train $K+2$ discriminators. Each discriminator $D_i$ outputs values
between [0, 1], representing the probability that $\T$ came from the data. In
practice we use the least square formulation \cite{LSGAN} for its
stability. Let $E$ represent the encoder including the image encoder and the 3D module. Then the adversarial loss function for the encoder is
\begin{equation}
 \label{eq:gan}
\min  L_{\text{adv}}(E) = \sum_i \mathbb{E}_{\T \sim p_{E}}[(D_i(E(I)) - 1)^2],
\end{equation}
and the objective for each discriminator is
\begin{equation}
  \label{eq:gan}
\min L_{\text{}}(D_i) = \mathbb{E}_{\T \sim p_{\text{data}}}[(D_i(\T) - 1)^2] +
  \mathbb{E}_{\T \sim p_{E}}[D_i(E(I))^2].
\end{equation}
We optimize $E$ and all $D_i$s jointly. 

%% file: src/detail.tex
\subsection{Implementation Details}
\label{sec:training}
\noindent {\bf Datasets:}
The \emph{in-the-wild} image datasets annotated with 2D keypoints that we use
are LSP, LSP-extended \cite{LSP} MPII \cite{andriluka14cvpr} and MS COCO 
\cite{coco}. We filter images that are too small or have less than 6 visible
keypoints and obtain training sets of sizes $1k$, $10k$, $20k$ and
$80k$ images respectively. We use the standard train/test split of these
datasets. All test results are obtained using the ground truth bounding box.

For the 3D datasets we use Human3.6M \cite{Human36m:2014} and 
MPI-INF-3DHP \cite{VNect}. We leave aside sequences from training Subject 8 of
MPI-INF-3DHP as the validation set to tune hyper-parameters, and use the full
training set for the final experiments. 
Both datasets are captured in a controlled environment and provide $~$150k training images with 3D joint annotations. For Human3.6M, we also obtain ground truth SMPL parameters for the training
images using MoSh \cite{Mosh} from the raw 3D MoCap markers. The unpaired data used to train the adversarial prior comes from
 MoShing three MoCap datasets: CMU \cite{cmumocap}, Human3.6M
training set \cite{Human36m:2014} and the PosePrior dataset
\cite{Akhter}, which contains an extensive variety of extreme
poses. These consist of 390k, 150k and 180k samples respectively.

All images are scaled to $224\times 224$ preserving the aspect ratio such
that the diagonal of the tight bounding box is roughly 150px (see \cite{LSP}).
The images are randomly scaled, translated, and flipped. Mini-batch size is 64. 
When paired 3D supervision is employed each mini-batch is balanced such that it
consists of half
2D and half 3D samples. All experiments use all datasets with paired 3D loss unless otherwise specified.

The definition of the $K=23$ joints in SMPL do not align perfectly with the
common joint definitions used by these datasets. We follow \cite{SMPLify,UP} and
use a regressor to obtain the 14 joints of Human3.6M from the reconstructed
mesh. In addition, we also incorporate the 5 face keypoints from the MS COCO dataset
\cite{coco}. New keypoints can easily be incorporated with the mesh
representation by specifying the corresponding vertex IDs\footnote{The vertex
  ids in 0-indexing are
nose: 333, left eye: 2801, right eye: 6261, left ear: 584, right ear: 4072.}. In total the
reprojection error is computed over $P=19$ keypoints. 

\noindent {\bf Architecture:}
\label{sec:architecture}
We use the ResNet-50 network \cite{He2016} for encoding the image,
pretrained on the ImageNet classification task \cite{tfslim}. The ResNet output is
average pooled, producing features $\phi \in \mathbb{R}^{2048}$. The 3D regression module
consists of two fully-connected layers with 1024 neurons each with
a dropout layer in between, followed by a final layer of 85D neurons. We use $T=3$ iterations for all of our experiments.
The discriminator for the shape is two fully-connected layers with 10, 5, and 1
neurons. For pose, $\myt$ is first converted to $K$ many $3\times 3$ rotation matrices
via the Rodrigues formula. Each rotation matrix is sent to a common embedding
network of two fully-connected layers with 32 hidden neurons. Then the outputs
are sent to $K=23$ different discriminators that output 1-D values. The
discriminator for overall pose distribution concatenates all
$K*32$ representations through another two fully-connected layers of 1024
neurons each and finally outputs a 1D value.  
All layers use ReLU activations except the final layer. 
The learning rates of the encoder and the discriminator network are set
to $1\times 10^{-5}$ and $1 \times 10^{-4}$ respectively. We use the Adam
solver \cite{Adam} and train for 55 epochs. Training on a single Titan 1080ti
GPU takes around 5 days. The $\lambda$s and other hyper-parameters are set
through validation data on MPI-INF-3DHP dataset. Implementation is in Tensorflow \cite{tf}.

%% file: src/experiments.tex
\section{Experimental Results}
\label{sec:exp}
\begin{figure*}[t]
  \centering
  \includegraphics[width=\textwidth]{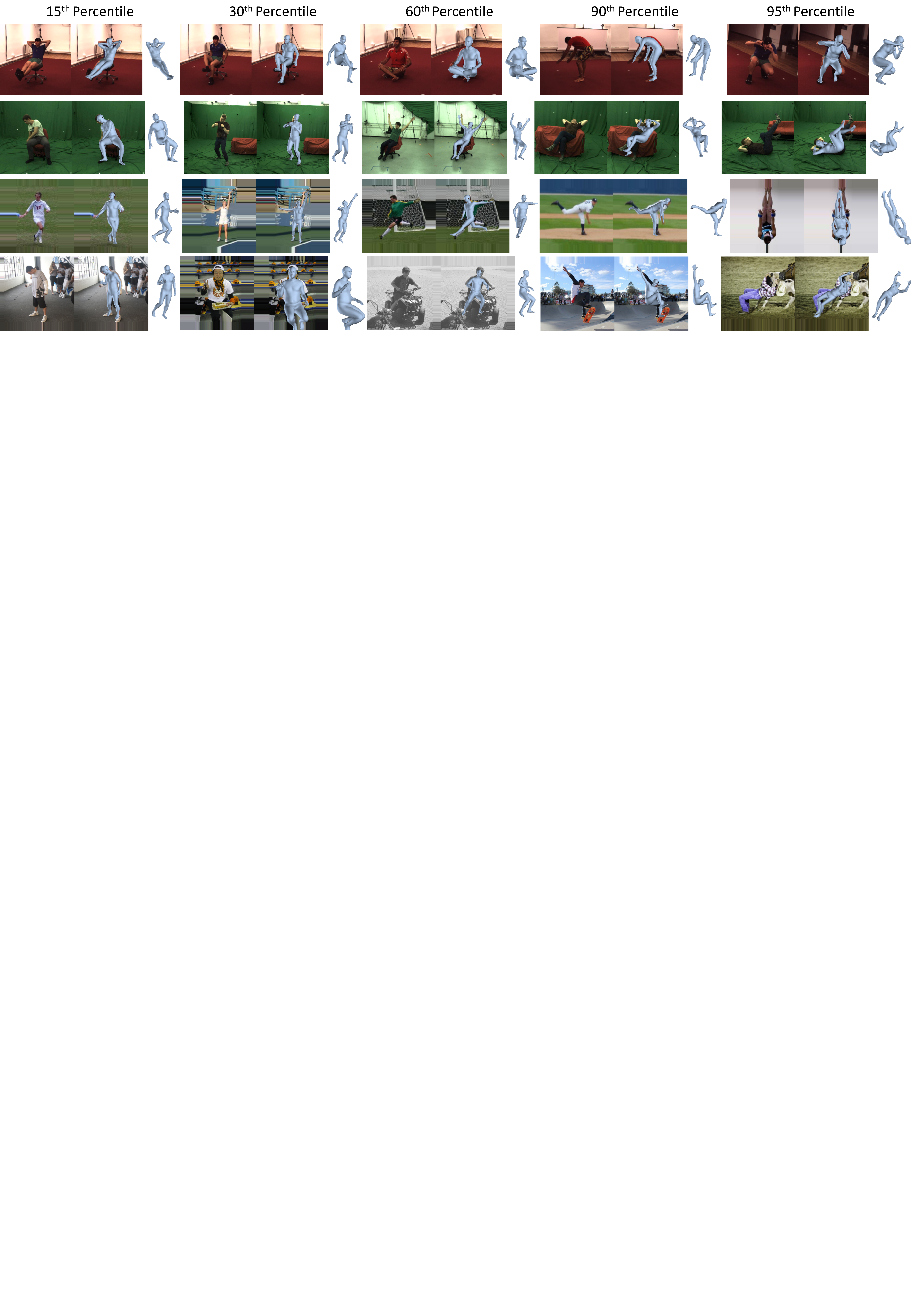}
  \caption{{\small {\bf Results sampled from different datasets at the 15th, 30th, 60th, 90th
    and 95th error percentiles.} Percentiles are computed using MPJPE for 3D
  datasets (first two rows - \blue{Human3.6M and MPI-INF-3DHP}) and 2D pose PCK
  for 2D datasets (last two rows - \blue{LSP and MS COCO}). \blue{High percentile
  indicates high error.} Note results at high error percentile are often semantically quite reasonable.}}
  \label{fig:bigfig}
\end{figure*}

Although we recover much more than 3D skeletons, evaluating the result
is difficult since no ground truth mesh 3D annotations exist for
current datasets. Consequently we evaluate quantitatively on the standard 3D joint 
estimation task. We also evaluate an auxiliary task of body part
segmentation. In Figure \ref{fig:teaser} we show qualitative results on
challenging images from MS COCO \cite{coco} with occlusion, clutter, truncation,
and complex poses. Note how our model recovers head and limb
orientations. In Figure \ref{fig:bigfig} \blue{we show results on
  the test set of Human3.6M, MPI-INF-3DHP, LSP and MS COCO} at various error
percentiles. Our approach recovers reasonable reconstructions even at 95th percentile error. Please see the project
website\footnote{https://akanazawa.github.io/hmr/} for more results. \blue{In
  all figures, results on the model trained with and without paired 2D-to-3D
supervision are rendered in light blue and light pink colors respectively.}

\subsection{3D Joint Location Estimation}
We evaluate 3D joint error on Human3.6M, a standard 3D pose benchmark captured in a
lab environment. We also compare with the more recent MPI-INF-3DHP \cite{Mehta17}, a dataset covering more poses
and actor appearances than Human3.6M. While the dataset is more
diverse, it is still far from the complexity and richness of in-the-wild images.

We report using several error metrics that are used for evaluating 3D
joint error. 
Most common evaluations report the mean per joint position error (\emph{MPJPE}) and \emph{Reconstruction} error, which is MPJPE
after rigid alignment of the prediction with ground truth via Procrustes Analysis
\cite{Gower1975}). Reconstruction error removes global misalignments and evaluates the quality of the reconstructed 3D skeleton.

\vspace{3mm}
\noindent\textbf{Human3.6M}
We evaluate on two common protocols. 
The first, denoted P1, is trained on 5
subjects (S1, S5, S6, S7, S8) and tested on 2 (S9, S11). Following previous work
\cite{Pavlakos,Rogez}, we downsample all videos from 50fps to 10fps to reduce
redundancy. The second protocol \cite{SMPLify,Tome_2017_CVPR}, P2,
uses the same train/test set, but is tested only
on the frontal camera (camera 3) and reports reconstruction error.

We compare results for our method (HMR) on P2 in Table \ref{tab:P2} with two recent approaches
\cite{SMPLify,UP} that also output SMPL parameters from a single
image. 
Both approaches require 2D keypoint detection as input and we out-perform both by a large
margin. We show results on P1 in Table \ref{tab:P1}. Here we also out-perform
the recent approach of Zhou \etal \cite{Xingyi2016}, which also outputs 3D joint
angles in a kinematic tree instead of joint positions.
Note that they specify the DoF of each joint by hand, while we learn this from data.
They also assume a fixed bone length while we solve for shape.
HMR is competitive with recent state-of-the-art methods that
only predict the 3D joint locations.

We note that MPJPE does not appear to correlate well with the visual quality
of the results.
We find that many results with high MPJPE appear quite reasonable as
shown in Figure \ref{fig:bigfig}, which shows results
at various error percentiles.

\begin{table}[t]
  \centering
\begin{tabular}{ll}
\hline
Method             & Reconst. Error \\ \hline
Rogez \etal \cite{RogezMocap}    & 87.3                \\
Pavlakos \etal \cite{Pavlakos}   & 51.9                \\
Martinez \etal \cite{Martinez2017}  & \textbf{47.7}                \\ \hline
*Regression Forest from 91 kps \cite{UP} & 93.9                \\
*SMPLify \cite{SMPLify}           & 82.3                \\
*SMPLify from 91 kps \cite{UP} & 80.7                \\ 
*HMR              & \textbf{56.8}                \\ \Xhline{.1\arrayrulewidth}% \hline
*HMR unpaired & 66.5                \\ \hline
\end{tabular}%
% }
 \caption{{\small {\bf Human3.6M, Protocol 2.} Showing reconstruction loss (mm);
     * indicates methods that output more than 3D joints. HMR, with and \emph{without} direct 3D
   supervision, out-performs previous approaches that output SMPL from 2D keypoints.}}
\label{tab:P2}
\vspace{-.5em}
\end{table}

\begin{table}[]
\centering
\resizebox{\columnwidth}{!}{%
\begin{tabular}{lcc}
\hline
Method                         & MPJPE & Reconst. Error \\ \hline
Tome \etal \cite{Tome_2017_CVPR}            & 88.39             &                     \\
Rogez \etal \cite{Rogez}                   & 87.7              & 71.6                \\
VNect \etal \cite{VNect}             & 80.5              &                     \\
Pavlakos \etal  \cite{Pavlakos}                & 71.9              & \textbf{51.23}               \\
Mehta \etal \cite{Mehta17}                 & 68.6              &                     \\
 % Zhou \etal \cite{Xingyi2017}  & 64.9              &                     \\
Sun \etal \cite{Xiao2017}                     & \textbf{59.1}              &                     \\ \hline
*Deep Kinematic Pose \cite{Xingyi2016} & 107.26            &                     \\
*HMR                          & \textbf{87.97}             & 58.1                \\ \hline
  *HMR unpaired  & 106.84 &  67.45              \\ \hline
\end{tabular}%
}
\caption{{\small {\bf Human3.6M, Protocol 1.}  MPJPE and reconstruction loss in mm.
 * indicates methods that output more than 3D joints.}}
\label{tab:P1}
\vspace{-1em}
\end{table}

\vspace{3mm}
\noindent\textbf{MPI-INF-3DHP}
The test set of MPI-INF-3DHP consists of 2929 valid frames from 6 subjects
performing 7 actions. This dataset is collected  indoors
and outdoors with a multi-camera marker-less MoCap system. Because of
this, the ground truth 3D annotations have some noise.
In addition to MPJPE, we report the Percentage of Correct Keypoints
(PCK) thresholded at 150mm and the Area Under the
Curve (AUC) over a range of PCK thresholds
\cite{Mehta17}. 

The results are shown in Table \ref{tab:mpiinf3dhp}. All methods use
the perspective correction of \cite{Mehta17}. 
We also report metrics after rigid alignment for HMR and VNect using the
publicly available code \cite{VNect}. We report VNect results without
post-processing optimization over time. Again, we are competitive with approaches that are trained to output 3D joints and
we improve upon VNect after rigid alignment.

\begin{table}[]
\centering
\resizebox{\columnwidth}{!}{%
\begin{tabular}{l|lll|lll}
  \hline
              & \multicolumn{3}{c|}{Absolute} & \multicolumn{3}{c}{After Rigid
                                                Alignment} \\ \hline
Method & PCK     & AUC     & MPJPE    & PCK          & AUC         & MPJPE        \\\hline
Mehta {\small\etal} \cite{Mehta17}        & 75.7    & 39.3    & \textbf{117.6}    & \multicolumn{1}{c}{-}            & \multicolumn{1}{c}{-}           & \multicolumn{1}{c}{-}            \\
VNect \cite{VNect}      & \textbf{76.6}    & \textbf{40.4}    & 124.7    & 83.9
                           & 47.3        & 98.0         \\ \hline
*HMR          & 72.9    & 36.5    & 124.2    & \textbf{86.3}         &
                                                                       \textbf{47.8}
                                      & \textbf{89.8} \\ \hline
*HMR unpaired     & 59.6    & 27.9    & 169.5    & 77.1  & 40.7 & 113.2 \\ \hline  
\end{tabular}%
}
\caption{{\small {\bf Results on MPI-INF-3DHP with and without rigid alignment.} 
* are methods that output more than 3D joints. Accuracy increases with alignment (PCK and AUC increase, while MPJPE
decreases).}}
\label{tab:mpiinf3dhp}
\end{table}

\subsection{Human Body Segmentation}
We also evaluate our approach on the auxiliary task of human body segmentation on the
1000 test images of LSP \cite{LSP} labeled by \cite{UP}. 
The images have labels for six body part segments and the background.
Note that LSP contains complex poses of people playing sports and
no ground truth 3D labels are available for training. We do not use the segmentation label during training either. 

We report the segmentation accuracy and average F1 score over all parts
including the background as done in \cite{UP}. 
We also report results on foreground-background segmentation. 
Note that the part definition segmentation of the SMPL mesh is not exactly the same as that of
annotation; this limits the best possible accuracy to be less than 100\%.

Results are shown in Table \ref{tab:seg}. 
Our results are comparable to the SMPLify oracle \cite{UP}, which uses ground truth segmentation and
keypoints as the optimization target. 
It also out-performs the Decision Forests of \cite{UP}. 
Note that \blue{HMR} is also real-time given a bounding box.

\begin{table}[]
\centering
\resizebox{\columnwidth}{!}{%
\begin{tabular}{l|ll|ll|l}
  \hline
  \multicolumn{1}{l|}{\multirow{2}{*}{Method}} & \multicolumn{2}{c|}{Fg vs Bg} &
                                                                                 \multicolumn{2}{c|}{Parts} & \multicolumn{1}{c}{\multirow{2}{*}{Run Time}} \\ % \cline{2-5}
\multicolumn{1}{l|}{}                         & Acc            & F1           & Acc          & F1          & \multicolumn{1}{c}{}                          \\ \hline

SMPLify \emph{oracle}\small{\cite{UP}} & 92.17         & 0.88         & 88.82        & 0.67        & -           \\
SMPLify \small{\cite{SMPLify}}     & 91.89         & 0.88         & 87.71        & 0.64        & $\sim$1 min \\
Decision Forests\small{\cite{UP}}& 86.60         & 0.80         & 82.32        & 0.51        & 0.13 sec    \\ \hline
HMR             & 91.67         & 0.87         & 87.12        & 0.60        &
                                                                               \textbf{0.04
                                                                               sec}
  \\ \hline
HMR unpaired            & 91.30         & 0.86         & 87.00        & 0.59        &
                                                                               \textbf{0.04 sec}    \\ \hline  
  
\end{tabular}%
}
\caption{{\small {\bf Foreground and part segmentation (6 parts + bg) on LSP \cite{UP}.} Reporting average accuracy and F1-score (higher the
  better). \blue{Proposed HMR} is comparable to the oracle SMPLify which uses ground truth
 segmentation in fitting SMPL.}}
\label{tab:seg}
\vspace{-1em}
\end{table}

\subsection{Without Paired 3D Supervision}
\label{sec:no3d}
So far we have used paired 2D-to-3D supervision, \ie $L_{\text{3D}}$ whenever
available. 
Here we evaluate a model trained without any paired 3D supervision. \blue{We refer to
this setting as \emph{HMR unpaired} and report numerical results in all the 
tables.}
All methods that report results on the 3D joint estimation task rely on direct 3D
supervision and cannot train without it. 
Even methods that are based on a reprojection loss
\cite{Pavlakos,Fish2017,3dinterpreter} require paired 2D-to-3D training data.

\begin{figure}[t]
  \centering
  \includegraphics[width=\linewidth]{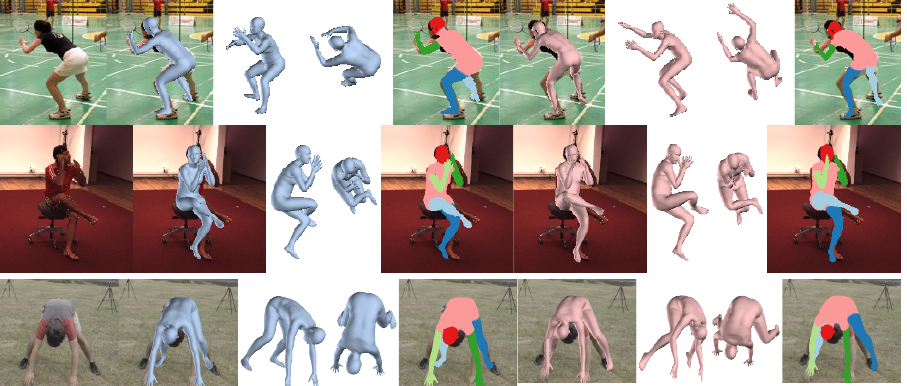}
  \caption{{\small {\bf Results with and without paired 3D
        supervision.} 
3D reconstructions, without direct 3D supervision, are very
      close to those of the supervised model.}}
  \label{fig:no3d}
  \vspace{-1em}
\end{figure}
 
The results are surprisingly competitive given this
challenging setting. 
Note that the adversarial prior is essential for training
without paired 2D-to-3D data. 
Figure \ref{fig:no3d} shows that a model trained with neither the paired 3D supervision
nor the adversarial loss produces monsters with extreme shape and
poses. 
It remains open whether  increasing the amount of 2D data will significantly increase 3D accuracy.  

\begin{figure}[t]
  \centering
  \includegraphics[width=\linewidth]{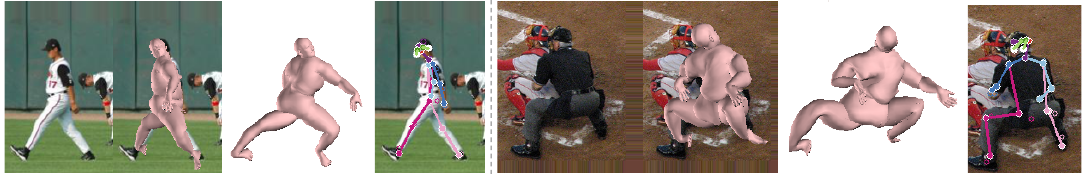}
  \caption{{\small {\bf No Discriminator No 3D.} With neither the
      discriminator, nor the direct 3D supervision, the network
    produces monsters. On the right of each example we visualize the ground
    truth keypoint annotation in unfilled circles, and the projection in filled
    circles. Note that despite the unnatural pose and shape, its 2D projection error is very accurate.}}
\label{fig:no3d}
\vspace{-1em}
\end{figure}

%% file: src/discussion.tex
\section{Conclusion}
\label{sec:discussion}
In this paper we present an end-to-end framework for recovering a full 3D mesh model of a
human body from a single RGB image. We parameterize the mesh in terms of 3D
joint angles and a low dimensional linear shape space, which has a variety of practical
applications. In this past few years there has been rapid progress in single-view 3D pose
prediction on images captured in a controlled environment. Although the performance on these benchmarks is starting to saturate, there
has not been much progress on 3D human reconstruction from images \emph{in-the-wild}. Our
results without using any paired 3D data are promising since they suggest that we
can keep on improving our model using more images with 2D labels,
which are relatively easy to acquire, instead of  ground truth 3D,
which is considerably more challenging to acquire in a natural setting.

\vspace{3mm}  
{\noindent {\bf Acknowledgements.}
We thank N.~Mahmood for the SMPL model fits to mocap
data and  the mesh retargeting for character animation, D.~Mehta for his assistance on MPI-INF-3DHP, and S.~Tulsiani,
A.~Kar, S.~Gupta, D.~Fouhey and Z.~Liu for helpful
discussions. This research was supported by BAIR sponsors and NSF Award IIS-1526234.}